\RequirePackage{fix-cm}
\documentclass[twocolumn]{svjour3}          
\smartqed  
\usepackage{graphicx}
 \journalname{Journal of Intelligent and Robotic Systems}

\usepackage{ifpdf}
\usepackage{cite}
\usepackage{graphicx}
\usepackage[cmex10]{amsmath}
\usepackage{algorithmic}
\usepackage{array}
\usepackage{stfloats}
\usepackage{fixltx2e}
\usepackage{stfloats}
\usepackage{url}
\usepackage{algorithm}
\usepackage{balance}
\usepackage{float}
\usepackage{color}
\usepackage[dvipsnames]{xcolor}
\newcommand {\mibf}[1] {\boldsymbol{#1}}

\usepackage{wrapfig}

\usepackage{hyperref}

\newcommand {\emilista} {\end{list}}

\usepackage[numbers]{natbib}

\newcommand {\Cs}{\hbox{{$\cal C$}}}                
\newcommand {\Cfree}{\hbox{{$\cal C_{\textrm{free}}$}}}                
\newcommand {\Cobstacle}{\hbox{{$\cal C_{\textrm{obs}}$}}}                
\newcommand {\Cint}{\hbox{{$\cal C_{\textrm{interaction}}$}}}                
\newcommand {\Cmove}{\hbox{{$\cal C_{\textrm{move}}$}}}                
\usepackage{lipsum}
\begin{document}\sloppy
%
\title{\textit{$\kappa$-PMP}: Enhancing Physics-based Motion Planners with Knowledge-based Reasoning \thanks{This work was partially supported by the Spanish Government through the projects 
\mbox{DPI2013-40882-P},  \mbox{DPI2014-57757-R} and  \mbox{DPI2016-80077-R}. Muhayyuddin is supported by the Generalitat de Catalunya 
through the grant FI-DGR 2014.
Aliakbar~Akbari is supported by the Spanish Government through the grant FPI 2015.}}
\author{Muhayyuddin,~
        Aliakbar~Akbari,~
        Jan~Rosell
\thanks{Muhayyuddin A. Akbari and J. Rosell are with Institute of Industrial and Control Engineering (IOC), Universitat Polit\`ecnica de Catalunya (UPC) -- Barcelona Tech\newline
\newline
\textbf{Muhayyuddin} \newline
muhayyuddin.gillani@upc.edu\newline
orcid.org/0000-0001-6214-1077\newline
\newline
\textbf{A. Akbari}\newline
aliakbar.akbari@upc.edu\newline
orcid.org/0000-0002-5290-9799\newline
\newline
\textbf{J. Rosell}\newline
jan.rosell@upc.edu\newline
orcid.org/0000-0003-4854-2370
}
}
\maketitle
\vspace{-20mm}
\begin{abstract}
Physics-based motion planning is a challenging task, since it requires the computation of the robot motions while allowing possible interactions with (some of) the obstacles in the environment. 
Kinodynamic motion planners equipped with a dynamic engine acting as state propagator are usually used for that purpose. The difficulties arise in the setting of the adequate forces for the 
interactions and because these interactions may change the pose of the manipulatable obstacles, thus either facilitating or preventing the finding of a solution path.  The use of knowledge can 
alleviate the stated difficulties. This paper proposes the use of an enhanced state propagator composed of a dynamic engine and a low-level geometric reasoning process that is used to determine how to 
interact with the objects, i.e. from where and with which forces. The proposal, called \hbox{\textit{$\kappa$-PMP}} can be used with any kinodynamic planner, thus giving rise to e.g. 
\hbox{\textit{$\kappa$-RRT}}. The approach also includes a preprocessing step that infers from a semantic abstract knowledge  described in terms of an ontology the  manipulation knowledge required by 
the reasoning process. The proposed approach has been validated with several examples involving an holonomic mobile robot, a robot with differential constraints and a serial manipulator, and 
benchmarked using several state-of-the art kinodynamic planners. The results showed a significant difference in the power consumption with respect to simple physics-based planning, an improvement in 
the success rate and in the quality of the solution paths. 

\end{abstract}

\section{Introduction}

Motion planning is one of the important issues in robotics, either as a stand-alone problem or in conjunction with other tasks such as grasping or manipulation. In the past decades, 
the focus of many motion planning approaches has often been to compute collision-free paths in the configuration space \Cs~(the set of all possible configurations of the robot) while satisfying some geometric constraints. Approaches like grid-based methods or potential fields  were first proposed for such planning~\cite{latombe1991}. Although being practical for simple scenarios, these algorithms were computationally intensive and difficult to implement in higher dimensional configuration spaces. Moreover, some problems arouse when  executing the computed geometric path in the real robot due to the possible existence of kinematic and dynamic constraints. Therefore, new motion planning algorithms appeared to cope with these problems~\cite{choset2005,lavalle2006}.

Sampling-based motion planning algorithms (such as RRT~\cite{lavalle2001}) were proposed to plan in higher dimensional configuration spaces, since these algorithms do not require the explicit 
representation of the obstacles in \Cs~and showed to comply very well to problems with kinodynamic constraints. In this line, the approaches such as Kinodynamic Motion Planning by Interior-Exterior Cell Exploration (KPIECE~\cite{sucan2012}) were recently proposed to plan efficiently for systems with complex dynamics in high dimensional configuration spaces. Moreover, these approaches can also take into account the physics-based constraints such as friction and gravity along with the kinodynamic constraints.

The possibility of taking into account physics-based constraints has also made it possible to consider not only the search of collision-free paths but also the search of paths where the interaction between the robot and the environment is possible. This new class of planning algorithms is known as physics-based motion planning,
and can be considered as an extension to kinodynamic motion planning. Physics-based motion planners evaluate the  interactions between bodies based on the principle of basic physics and their results influence the planning process. Therefore, planning becomes a challenging task due to various factors such as the high dimensionality of the state space (where the environment may change as a result of the bodies interactions), the large planning search space and the possibly highly constrained solution set. Moreover, the accurate modeling of the objects interactions with dynamical effects such as friction and momentum is required. To alleviate these challenging issues, a few approaches have been proposed that develop strategies to reduce the search space, e.g.~\cite{zickler2009, zickler2010, muhayyuddin2015}.

Since physics-based motion planning deals with the kinodynamic and physics-based
constraints along with the dynamic interactions between rigid bodies, it is desirable to compute an efficient solution in terms of dynamics measures, such as power consumption, instead of just considering the planning time or path length, as usually done in other motion planning problems (this is particularly true for task and motion planning problems where dynamic cost computed by the motion planner significantly influence the planning decision at task level~\cite{Ali2015}~\cite{RobotAli2015}). To the best of our knowledge, there is not any approach within the framework of physics-based planning that searches for power-efficient motion plans. 

With the physics-based motion planning in mind, one of the problems that can be stated is the following. Find a path for a robot (either a mobile or a manipulator) to move from an initial 
state to a goal state while interacting, if necessary, with the obstacles of the environment,  obtaining a power-efficient solution that satisfies the constraints (regarding which obstacles can be 
collided, from where and with which interacting force).
This paper aims to find a solution to this problem statement by proposing a power-efficient approach for physics-based motion planning based on the use of manipulation knowledge.
Compared with plain physics-based motion planners, the proposal results in an improvement in the success rate  and in the quality of the solution paths. A preliminary version of this approach was presented in~\cite{muhayyuddin2015}, where the focus was on the introduction of a knowledge-based reasoning process for an efficient planning (in terms of time). This paper extends the  approach for a power-efficient solution by proposing an improved framework and a new planning algorithm. Moreover, a detailed comparison of the approach with other approaches is presented in a variety of different scenarios.

\textit{Contributions:}
The main contribution of this work is the proposal of a physics-based motion planning method equipped with a control sampling strategy that allows the search of a  power-efficient motion plan, which 
may include free robot motions along with interactions with manipulatable objects that may be obstructing the path.
  The main components of the proposal are:
  \textit{(1)} the partition of \Cs~into different regions as a function of whether the robot can either move freely or interact with manipulatable objects; \textit{(2)} an instantaneous reasoning 
process that performs low-level geometric reasoning in order to analyze the configuration space regions and update the sampling range;
   \textit{(3)} a detailed representation of knowledge, which is categorized into semantic knowledge and manipulation knowledge, to help improving the knowledge-based inference process.
 The proposed framework consists of a high-level and a low-level layer, that are connected through a ROS-based communication layer. The high-level layer contains the knowledge about the 
robot and the environment, that is used by the low-level motion planner for reasoning about the sampled controls and to update the manipulation constraints. This hierarchical structure results in 
smart motion plans for the robot to efficiently interact with the objects in the environment.
The performance of the proposed planning approach is tested with three different scenarios: an holonomic mobile robot, a car-like mobile robot and a planar manipulator. The results are compared (in 
terms of power consumption, planning time and success rate) with other physics-based planning approaches.

The paper is structured as follows. First, Sec.~\ref{s-relatedwork} details some relevant related work and Sec.~\ref{s-overview} formulates the problem and sketches the solution. Afterwards, 
Sec.~\ref{s-Knowledgerepresentation} explains the high and low level representation of knowledge, Sec.~\ref{s-Knowledgeinference} details the high level knowledge inference process and the low level 
reasoning process and Sec.~\ref{s-imlementation} explains the framework and the proposed planning algorithm. Finally,  Sec.~\ref{s-resultsdiscussion} describes the obtained results and 
Sec.~\ref{s-conclusion} concludes the study.
\section{Related Work}\label{s-relatedwork}

The simplest form of motion planning is a geometric problem  devoted to compute a collision-free path from a start to a goal state in the configuration space while satisfying some geometric 
constraints like joint limits and collision avoidance. Sampling-based motion planners such as RRTs and PRMs~\cite{kavraki1996} are able to solve problems in high dimensional configuration spaces, by 
connecting collision-free samples forming a graph or a tree-like structure that capture the connectivity of the free configuration space, or of the part of the free space relevant to the query to be 
solved. In some cases the kinematic and dynamic constraints of the robot must be taken into account while planning due to the difficulty that may arise in the following of a geometric path. This need 
gave rise to kinodynamic motion planners.

\subsection{Kinodynamic Motion Planning}
Physical systems may be subject to kinematic constraints that may be holonomic or nonholonomic. The former are 
constraints on system configurations, $q_0, q_1, . . ., q_n$, and can be expressed as \hbox{$f (q_0, q_1, . . ., q_n; t) = 0$}, 
i.e. they only depend on the coordinates and time, whereas the latter cannot be expressed in this form; they are constraints on velocities. Moreover, the system may be subject to dynamic constraints 
due to the dynamics laws and bounds on velocities, accelerations and applied forces.
In robotics, the term kinodynamic planning was introduced by~[10] to refer to motion planning problems where both kinematic and dynamic constraints were considered, i.e. a motion planning  approach 
devoted to the search of a solution path that complies to the kinematic and dynamic laws and the bounds over the applied forces, the velocities and the accelerations.
Basically, these kinodynamic algorithms search a solution in a higher dimensional state space $\cal X$ that records the dynamics of the system. For any particular configuration $q \in $ \Cs, the state 
of the system is represented as $x= (q,\dot{q})\in \cal X$, and the state propagation is performed by a transition function defined as:
\begin{equation}\label{eq:Tfunction}
 \dot{x}=f(x,u)
\end{equation}
with $u\in U$, the  set of valid control inputs. The solution to a kinodynamic problem is found by determining the set of appropriate control inputs that applied using Eq.~(\ref{eq:Tfunction}) brings 
the robot from the initial to the goal state while satisfying all the constraints.

 Sampling-based motion planners (particularly those using tree-like structures) have the ability to  efficiently plan in the presence of kinodynamic constraints~\cite{carpin2006}\cite{csucan2010}. 
These planners can be divided into three main categories:
 \begin{enumerate}
 \item Planners that sample the states, such as RRTs and Expansive-Spaces Tree planners (EST)~\cite{hsu2002}~\cite{hsu1997}.
 The RRT grows a tree rooted at the start state by iteratively selecting a random sample $x_{rand}$ and expanding the tree from the node that is nearest to  $x_{rand}$ by applying a randomly  sampled 
control. The EST builds a tree-like roadmap by selecting a
 node with a probability inversely proportional to the density of the node neighborhood and extending it by applying a randomly sampled control.

 \item Planners that sample path segments instead of states and that do not require the use of a metrics, such as the Path Directed Subdivision 
Tree (PDST)~\cite{ladd20051} and 
KPIECE~\cite{csucan2010}. The PDST planner subdivides the state space into regions (cells), each one containing a path segment. The tree is extended by iteratively sampling a cell, randomly selecting 
a state from that cell, and applying a randomly sampled control to generate the new path-segment. 
\\The KPIECE planner also samples the path segments (called motions) by using a grid-based decomposition 
of a projection space (defined either with random projections or user-defined) where the tree of motions is projected and where a sampling procedure is defined to select the important regions to 
explore.

 \item Hybrid planners such as Synergistic Combination of Layers of Planning (SyCLoP)~\cite{Plaku2010} and the Linear Temporal Logic (LTL) motion planner~\cite{bhatia2011}~\cite{plaku2012}. The SyCLoP 
planner splits the planning problem into a discrete (high-level) layer and a continuous (low-level) layer of planning. The former is based on the decomposition of the workspace, whereas the latter 
consists of a sampling-based motion planner like EST or RRT that is guided by the discrete layer. LTL is an extension of the SyCLoP planer in which the discrete layer encodes
     a complex motion planning task using an abstract graph computed from a decomposition of the workspace and an automaton that represents a linear temporal logic formula describing the task.
\end{enumerate}

 In all the above stated planners the control sampling range is usually set at the start and remains the same in the entire planning process; on each state the controls are randomly sampled from the 
given control range that results in the robot motion. Beside sampling-based algorithms there are some other recently proposed approaches for kinodynamic motion planning, such as the Covariant 
Hamiltonian Optimization for Motion Planning (CHOMP~\cite{zucker2013}). These approaches mainly focus on the optimization (such as smoothness) but can be used as stand alone motion planners for 
computing collision-free trajectories. 

\subsection{Physics-Based Motion Planning}

All the above stated kinodynamic motion planning strategies are focused on computing a collision-free trajectory from start to the goal state, i.e. interactions with the objects in the environment are forbidden (and the contacts of a mobile robot with the floor not considered). This constraint leads to the neglecting of the physics-based dynamic features such as friction between the objects and the ground, pressure distribution under the objects surfaces, gravitational effect over the objects in the environment and the interaction dynamics (such as direction of interaction force and momentum). 
If interactions are allowed, however, all these features should be considered, and this is what physics-based motion planning does, and the key difference with respect to the kinodynamic and other 
motion planning approaches. Therefore, physics-based motion planning has recently emerged as a step further towards physical realism. 
It simultaneously considers the kinodynamic constraints and physics-based constraints (such as friction 
and gravity), and also incorporates the purposeful manipulation of objects by evaluating the dynamic interactions between rigid bodies  simulated using the basic physics (rigid-body dynamics).

Physics-based  planning approaches implicitly use a sampling-based kinodynamic planner, that is responsible for sampling the states and constructing the solution path, but using for the propagation step a 
rigid-body dynamic simulator (dynamic engine), such as ODE~\cite{OpenDE2007} or Bullet~\cite{bullet2013}. The dynamic engine models the dynamical world with all the physical properties and has the 
ability to simulate the properties of the physical interactions, such as force-based inter-body collision and momentum. The physics-based planner evaluates the results of the action after propagation, 
if it satisfies all the constraints then the action is selected, and discarded otherwise. 

The complexity of the physics-based motion planning is very high due to the high-dimensional state space, large search space and highly constrained solution set. A few physics-based motion planning 
approaches have been proposed that addressed the above mentioned issues, such as the 
Behavioral Kinodynamic Balanced Growth Trees (BK-BGT) and  the Behavioral Kinodynamic
Rapidly-Exploring Random Trees (BK-RRT) proposed in~\cite{zickler2009} that use a strategy to reduce the search space  based on a nondeterministic tactic modeled using a finite state machine, along 
with skills used to control the sampling. The propagation step is performed using PhysX~\cite{physx}. A hybrid approach is proposed in~\cite{muhayyuddin2015} that equips the physics-based motion 
planner with knowledge (in the form of ontologies) about the robot's manipulation world. It uses a knowledge-based reasoning process to reduce the robot search space and guide the motion planner by 
defining the way objects can be manipulated. It uses RRT and KPIECE as kinodynamic motion planner and ODE as state propagator. This approach has also been used in task planning 
approaches~\cite{Ali2015}~\cite{RobotAli2015} for the physics-based reasoning process to determine the feasibility of the plan by evaluating the dynamic cost of each subaction in the task plan.

Some other approaches address problem related to physics-based motion planning such as physics-based grasping and rearrangement planning~\cite{dogar2011, dogar2012}. These 
approaches evaluate the dynamic interaction by executing the straight line trajectories under the quasi static assumption. Moreover some approaches (such as~\cite{Haustein2015,stilman2005}) studied 
the rearrangement planing in conjunction with the physics-based motion planning, but none of them addressed the issue of robust control selection for the power efficient solution.

The determination of the appropriate set of controls and durations such that, if sequentially applied, the robot  moves  from an initial to a goal state following a power-efficient 
trajectory is a challenge. With this aim,  the current proposal presents a control sampler that uses a knowledge-based reasoning process to determine the appropriate values of controls that may result 
in this behavior. The next section will formally describe the problem statement and outline the proposed solution.

\begin{figure}[t]
\begin{center}
   \includegraphics[width=0.8 \linewidth]{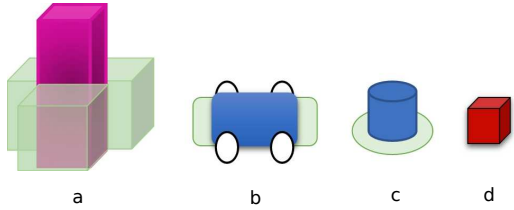}
   \caption{Different bodies with their \textit{mRegions} (the red cube represents the fixed object and hence have no associated \textit{mRegion}).} \label{fig:objects}
\end{center}
\end{figure}
\begin{figure}[t]
\begin{center}
   \includegraphics[width=0.85 \linewidth]{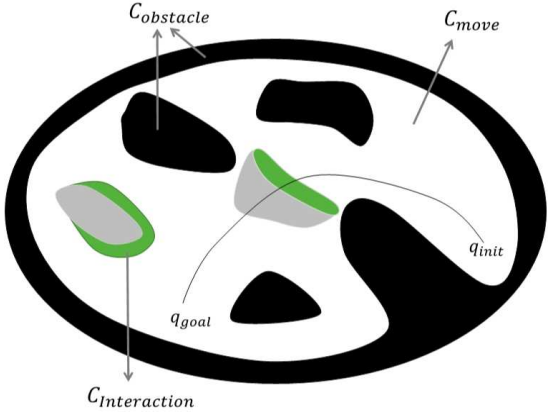}
   \caption {Configuration space: \Cobstacle\ shown in black (collisions with fixed obstacles) and gray (collision with manipulatable obstacles), \Cint\ in green and \Cmove\ in white.}\label{cspace}
\end{center}
\end{figure}
\section{Problem Formulation and Solution Overview}\label{s-overview}

\subsection{Problem Statement}
Let a physics-based motion planning problem be defined as the tuple $(\mathcal{X}, \mathcal{U}, f,\mathcal{K}, \mathcal{F}, x_{init}, \mathcal{X}_{goal})$, where:
\begin{itemize}
\item $\mathcal{X}$ represents the state space; it is a differential manifold.
\item $\mathcal{U}$ represents the control space; it contains the set of all possible control inputs that can be applied to the robot.
\item $f:\mathcal{X}^i\times\mathcal{U}\longrightarrow\mathcal{X}^{i+1}$ is the propagation function.
\item $\mathcal{K}$ is the abstract knowledge containing all the available knowledge about the world such as object classification, manipulation regions, and physical properties. $\kappa \subset 
\mathcal{K}$ is the instantiated knowledge that represents the knowledge that is valid for a particular instance of time.
\item $\mathcal{F}:\kappa\times\mathcal{X}\longrightarrow\{0,1\}$ is the physics-based state validity checker. It evaluates the state generated by applying $f$, and returns 1 if it satisfies all the 
constraints imposed by~$\kappa$, or returns 0 otherwise.
\item $x_{init} \in \mathcal{X}$ is the initial state.
\item $\mathcal{X}_{goal} \subset \mathcal{X}$ is the goal region.
\end{itemize}

Consider a motion planning problem where no collision free trajectory from start to the goal exist (e.g. either the goal or the way towards it might be occupied with manipulatable objects). The 
objective is to determine the set of efficient control inputs $\{u_1,\dots,u_n\} \in \mathcal{U}$ and the set of associated time durations $\{t_1,\dots,t_n\}$ such that, if sequentially applied to 
the system using $f$ and starting from $x_{init}$, a goal state $x_n \in \mathcal{X}_{goal}$ is reached, being the resultant trajectory power-efficient and satisfying all the constraints (i.e. a 
trajectory that avoids collisions with fixed obstacles but that may collide with manipulatable objects to push them away from the solution path).
The proposed approach (at each step) will
determine the robust control using the low level reasoning about the object dynamics, in such a way that the resultant solution will be power efficient (no optimization of any kind such as path 
length will be considered).

\subsection{Problem modeling}

We consider a dynamical workspace with several objects (each one composed of one or more rigid
bodies), that are categorized into \textit{fixed objects} and \textit{manipulatable objects}.
Fixed objects, such as walls, remain static during the entire planning process and no collision is allowed with them.  Manipulatable objects, on the contrary, can be pushed and hence their pose is 
not fixed. Manipulatable objects are further categorized into \textit{constraint-oriented manipulatable objects} (\textit{co-mObjects}) and \textit{free-manipulatable objects} 
(\textit{free-mObjects}), depending on whether they are subject to some kinematic constraints or not (e.g. a car-like object can only be pushed forwards or backwards in a single direction). 
All manipulatable objects have associated regions, called manipulation regions (\textit{mRegions}), from where the robot can exert forces in order to move them, i.e. the robot can  interact with the object 
through these regions and it is not allowed to contact the object from any other part. For instance, as shown in Fig.~\ref{fig:objects}, a car-like robot has one  \textit{mRegion} located at its front 
and one at its rear, and a vertical box has the \textit{mRegions} around it but below its center of mass.

The state $s$ of each object is represented by its position $r$, orientation $o$, linear velocity $v$, angular velocity $w$, and the associated manipulation constraints $\eta$:
\begin{equation}
 s=\{r,o,v,w,\eta\}
\end{equation}
The state of the workspace (composed of $n$ objects) at a given instant of time $t$ is represented as:
\begin{equation}
 \mathbf{q}_t=\{s_1,s_2\dots,s_n,t\}
\end{equation}

Regarding the robot, its configuration space \Cs, i.e. the set of all possible configurations of the
robot, can be divided into the set of geometrically accessible regions, called \Cfree,~and the set of the forbidden ones (those corresponding to collision with obstacles), called \Cobstacle. The 
condition \Cfree $\cup$ \Cobstacle $=$ \Cs\ holds. In this work,
\Cfree~is further divided into~\Cmove~and~\Cint\ representing, respectively, the regions where the robot can move freely and those where the robot can enter in contact with the manipulatable objects, 
i.e. the set of configurations corresponding to the robot\footnote{In the case of manipulators the pose of the tool will be considered.} being placed in an \textit{mRegion}  (see Fig.~\ref{cspace}). 
The condition~\hbox{\Cmove $\cup$ \Cint $=$ \Cfree} holds.


\subsection{Solution Overview}
In order to find a power-efficient solution, we have developed an approach inspired by our daily life experience where, in order to perform the task robustly,  we dynamically update the forces according to the interaction with the environment, e.g. we do not exert the same force while pushing a plate, a table, or when moving freely. We propose the use of any kinodynamic planner, such as KPIECE or RRT, and to equip it with a state transition model that computes the next state based on a dynamic engine and a reasoning process that uses instantiated knowledge. This knowledge defines from where objects should be manipulated and with which range of forces.
 The knowledge representation used is detailed in Section~\ref{s-Knowledgerepresentation} and the knowledge inference and reasoning process in Section~\ref{s-Knowledgeinference}.
 
\begin{figure}[t]
\begin{center}
   \includegraphics[width=1 \linewidth]{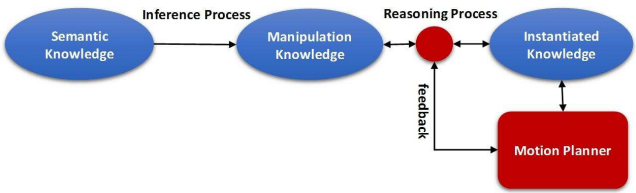}
   \caption{Flow of knowledge from the high-level abstract knowledge to the low-level instantiated knowledge.}\label{fig:knowflow1}
\end{center}
\end{figure}

The knowledge about the task and the workspace is modeled in two levels:
\begin{itemize}
\item The \textit{abstract knowledge} $\mathcal{K}$: It is a high-level representation of knowledge composed of the \textit{semantic knowledge} $\mathcal{K}_\textrm{S}$ and the  \textit{manipulation 
knowledge} $\mathcal{K}_\textrm{M}$. The semantic knowledge describes, using ontologies, information of the task such as the kinematic and dynamic properties of the robot, of the objects, and the manipulation constraints. From $\mathcal{K}_\textrm{S}$ the manipulation knowledge $\mathcal{K}_\textrm{M}$ is inferred. It contains all the necessary information that is required for the motion planner, such as the type of objects and how they can be manipulated. Abstract knowledge remains the same during the whole planning process.
\begin{figure}[t]
\begin{center}
   \includegraphics[width=0.88\linewidth]{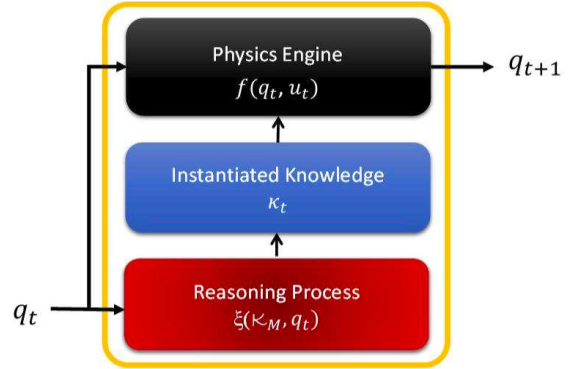}
   \caption{State transition model for the \textit{$\kappa$-PMP}.}\label{fig:propagator}
\end{center}
\end{figure}
\item The \textit{instantiated knowledge} $\kappa$: It is a low-level representation, updated at each instance of time by the reasoning process, based on the manipulation knowledge and on the feedback received from the motion planner (e.g. if at a particular instance of time one of the \textit{mRegion} of an \textit{mObject} is occupied by some other object, then if there is an \textit{mRegion} in the opposite side, it will be deactivated by the reasoning process and $\kappa$ will be updated accordingly). The instantiated knowledge is used in the state transition model as explained next.
\end{itemize}

The dynamic flow of knowledge is shown in Fig.~\ref{fig:knowflow1}, where it is illustrated that the manipulation knowledge is extracted from semantic knowledge (ontologies) and  used  by the 
reasoning process along with the feedback from the motion planner to update the instantiated knowledge. This instantiated knowledge, as shown in Fig.~\ref{fig:propagator}, is the key module of the 
state transition process, that takes the state of the world $\mathbf{q}_t$ as input and generates the next state $\mathbf{q}_{t+1}$ by applying an appropriate control
based on this knowledge:

 \begin{eqnarray}\label{eq:state propagator}
	 \kappa_{t} &=& \xi(\mathcal{K}_M, \mathbf{q}_{t})\\
	\mathbf{q}_{t+1} &=& f(\mathbf{q}_t,\mathbf{u}_t(\kappa_{t}))
\end{eqnarray}
That is, the state transition process consists of three modules: the 
physics engine, the instantiated knowledge, and the low-level reasoning. The reasoning process 
is responsible for updating the instantiated knowledge $\kappa$ with the manipulation constraints (determining which are the active manipulation regions) and with the region of the configuration space 
where the robot is located (either \Cmove\ or \Cint). With this information the instantiated knowledge determines the set of controls  and selects one to be applied (the set will be determined in such 
a way that if the robot is in a region where interaction with an object is possible then the range of controls must allow the manipulation of the object). Finally, the physics engine (i.e. 
physics-based state propagator) generates $\mathbf{q}_{t+1}$ by applying the randomly sampled control $\mathbf{u}_t$  to $\mathbf{q}_t$. The validity of the resulting state will be checked using the  
physics-based state validity checker $\mathcal{F}$ that takes into account the instantiated knowledge (i.e. a collision state is only valid if the robot collides with manipulatable object from one of 
its manipulation regions).

 The preliminary work in \cite{muhayyuddin2015} demonstrated the importance of using knowledge (described by ontologies) to guide physics-based motion planning. The  present proposal greatly enhances 
the state transition model, which is the core of the proposal, by introducing the low-level geometric reasoning process that allows the use of an adaptive control range, thus obtaining power-efficient 
solutions. On the other hand, it reduces the computational cost by introducing the offline inference process that prevents the online query to the ontological knowledge.

 
 \begin{figure}[t]
\begin{center}
   \includegraphics[width=1.0 \linewidth]{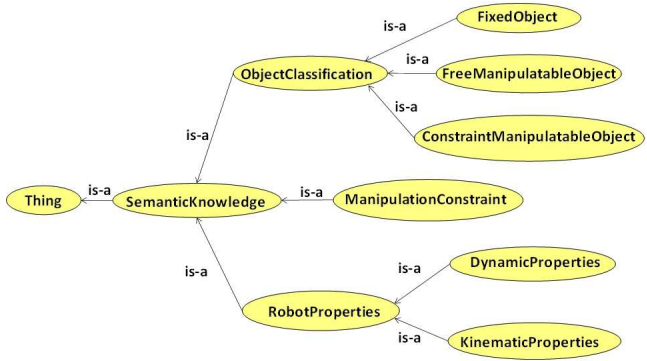}
   \caption{OWL-based semantic knowledge taxonomy. 
https://sir.upc.edu/projects/ontologies.}\label{fig:texonomy}
\end{center}
\end{figure}
\section{Knowledge Representation}\label{s-Knowledgerepresentation}

   This section presents the proposal done to represent and manage knowledge for motion planning purposes, from the abstract knowledge $\mathcal{K}_S$ using ontologies to the instantiated knowledge $\kappa$ used at the motion planning level, through the manipulation knowledge  $\mathcal{K}_M$ inferred from  $\mathcal{K}_S$  and used to maintain  $\kappa$.
    
\subsection{ Representation of Knowledge using Ontologies}

Ontologies can be employed to model and organize knowledge within different domains.
In particular, they have  recently been used
to organize knowledge for motion and manipulation planning (e.g. \cite{feyzabadi2014}\cite{Ali2015}) in order to enhance inference capabilities.
 Ontologies can be easily edited using the \textit{Prot\'eg\'e} editor~\cite{protege} and can be encoded and stored using the Web Ontology Language (OWL)~\cite{owl2004}. In this way, the knowledge can 
be shared by different devices, being accessed through the world wide web.
  Using ontologies, knowledge is mainly classified in classes (which entails a collection of objects), individuals (in which instances of classes are stored), relations (expressing the correspondence 
among objects as well as individuals), and properties (specifying data values for objects).

\subsection{Abstract Knowledge}

The abstract knowledge is the high-level representation of knowledge which remains fixed throughout the planning process. It is divided into the \textit{Semantic Knowledge} containing information 
about the workspace and the robot, coded as an OWL taxonomy, as depicted in Fig.~\ref{fig:texonomy}, and the \textit{Manipulation Knowledge} which involves knowledge related to how the robot can 
interact with the workspace, and which is inferred from the semantic knowledge. 

\begin{figure}[h]
\begin{center}
   \includegraphics[width=1.0 \linewidth]{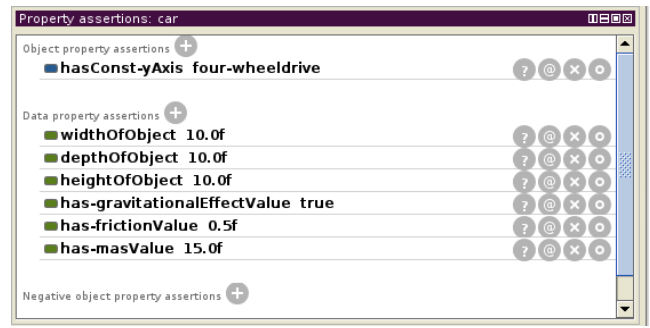}
   \caption{Screen shot of the Prot\'eg\'e editor showing the semantic properties of a car-like object.}\label{fig:semanticprop}
\end{center}
\end{figure}

\subsubsection{Semantic Knowledge ($\mathcal{K}_S$)}\label{sec:semKnowledge}
 Semantic knowledge categorizes information within the following classes:

\begin{itemize}
  \item Class \textit{``RobotProperties''} describes the properties of the robot in two subclasses. Geometric constraints of the robot such as joint limits are stored in the class 
\textit{KinematicProperties}; differential properties of the robot such as bounds on forces, torques, velocities, and accelerations (global properties that condition the maximum capacity of the robot) 
are stored in the class \textit{DynamicProperties}.
  \item Class \textit{``ObjectClassification''} is used to describe the objects in the workspace such as, fixed 
(\textit{fixedObject}), free manipulatable (\textit{free-mObject}) and constraint-oriented manipulatable (\textit{co-mObject}).

  \item Class \textit{``ManipulationConstraint''} describes orientation constraints on the motion of bodies and objects.
 
\end{itemize}

The hierarchy of knowledge among the classes can be represented using Description Logic (DL). For instance the hierarchy for a constraint-oriented manipulatable object is described below:\\
\\
$Thing:= object$\\ 
 $  \exists hasSuperclass (Thing, SemanticKnowledge)$\\
 $ \wedge \exists hasClass (SemanticKnowledge, ObjClassification)$\\
 $ \wedge \exists hasSubclass(ObjectClassification, co\textrm{-}mObject)$\\ 
 
 where $\wedge$ and $\exists$ represents \textit{conjunction} and \textit{exist}, respectively.

 The semantic properties (that are stored on OWL) are divided into object properties and data properties. The former are used to describe the relationships between the individuals, and the latter are used to assign the values to the physical attributes.  As an example, some of the semantic properties of the car-like object are depicted in  
Fig.~\ref{fig:semanticprop} and explained below in terms of DL.\\ 
\\
 $ Object:= Car$\\
 $ \wedge \exists hasWheel (Car, Wheel)$\\
 $ \wedge \exists hasBody (Car, Body)$\\
 $ \wedge \exists hasWheel (Wheel, alongYaxis)$\\
 $ \wedge \exists canMove (Car, alongXaxis)$\\

The above stated DL description of an object property (\textit{hasConst-yAxis fourwheeldrive}) explain that car is composed of wheels and body, the motion of the car-like object is constraint by the wheels that are along the y axis of the chassis so that the car can only move along the associated x-axis. 

The data properties in terms of description logic are represented as follows:\\
\\
 $Object:= Car$\\
 $  \exists hasWidth (Car, Value)$\\
 $ \wedge \exists hasDepth (Car, Value)$\\
 $ \wedge \exists hasHeight (Car, Value)$\\
 $ \wedge \exists hasGravity (Car, Value)$\\
 $ \wedge \exists hasFriction (Car, Value)$\\
 $ \wedge \exists hasMass (Car, Value)$\\

It describe the dimension of the car, response to the gravitational (if considered as dynamic object, the value will be true and false otherwise) and the values of 
friction (between wheels and road) and mass of the car respectively.

\subsubsection{Manipulation Knowledge ($\mathcal{K}_M$) }

Manipulation knowledge is inferred from $\mathcal{K}_S$  using a Prolog inference process that will be detailed later in Sec.~\ref{sec:prologinf}. It contains the classification of 
objects 
(\textit{fixedObject}, \textit{free-mObject}, or 
\textit{co-mObject}), a complete set of \textit{mRegions} for manipulatable objects, physical attributes of objects, and kinematic and dynamic properties of the robot (such as joint limits and bounds 
on forces and velocities). Manipulation knowledge remains fixed throughout the motion planning process. $\mathcal{K}_S$ contains the maximum possible knowledge about the world and $\mathcal{K}_M$ 
that related to a particular motion planning problem (e.g. it must be updated if the features of the robot change).

\subsection{Instantiated Knowledge ($\kappa$)}
Instantiated knowledge is the low level representation of  knowledge (at the motion planning level). It is dynamic and valid for a particular instant of time (containing all possible constraints that are valid for that instant of time) and is updated for the next time step. The instantiated knowledge contains two main components: \textit{(1)} the knowledge about the valid and invalid manipulation regions, as well as  dynamical properties (such as masses, friction coefficients) of the objects; \textit{(2)} the control sampling range (such as bounds on control forces and torques) to be used by the motion planner at the current instant of time (by using the state propagator, i.e. the dynamic engine).

The instantiated knowledge is updated by the low-level reasoning process  based on the high-level manipulation knowledge and the feedback from the motion planner. The reasoning process is detailed in Sec.~\ref{sec:reasoning}).

\section{Knowledge Inference and Reasoning Process}\label{s-Knowledgeinference}

The flow of knowledge was graphically illustrated in Fig.~\ref{fig:knowflow1}. It includes the knowledge inference process and the reasoning process.
The knowledge inference process is the pre-processing step responsible of inferring the manipulation knowledge~$\mathcal{K}_M$ from the ontological semantic knowledge~$\mathcal{K}_S$. On the other hand, the reasoning process is the responsible of updating the instantiated knowledge at each instant of time while planning, based on $\mathcal{K}_M$ and the current state fed back by the motion planner.

\subsection{Knowledge Inference Process}\label{sec:prologinf}

The inference process is performed using Knowrob~\cite{knowrob2009}, a knowledge-based processing tool for robotics applications that allows operating over OWL data bases, extended with the following predicates (coded in Prolog) tailored to extract the necessary information for the physics-based motion planner:
\begin{itemize}
\item \textit{object\_classification(?Obj, ?ObjType)}: Given an object \textit{Obj} returns the type of the associated object \textit{ObjType} by evaluating its category. Those manipulatable objects 
that are too heavy for the robot to be manipulated are changed to fixed.

\item \textit{manipulatable\_region(?Obj, ?ManipRgns)}: Given a manipulatable object \textit{Obj} computes the set of  associated manipulatable regions \textit{ManipRgns} taking into account the 
manipulation constraints, if any.
\item \textit{object\_properties(?Obj, ?PhysicalProps, ?Dimension)}: Given an object \textit{Obj} returns its physical properties \textit{PhysicalProps}, including mass and friction values, as well as 
gravitational effect  and the dimension \textit{Dimension} of the object.

\item \textit{robot\_properties(?DynamicsProps, ?KinematicProps)}: Returns the dynamics properties of the robot (forces and velocities limits) in \textit{DynamicsProps}, and the kinematic properties 
(joint limits) in \textit{KinematicProps}.
\end{itemize}
\begin{figure}[t]
\begin{center}
   \includegraphics[width=1\linewidth]{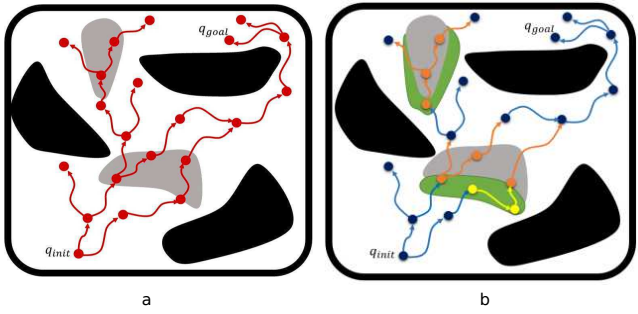}
   \caption{Sample and propagation using: (a) the simple
physics-based motion planner; (b) $\kappa$-PMP. The control sampling range is selected based on the dynamics properties of the target object and the manipulation
region. Blue samples represent the lower values of control range, whereas orange samples represent the higher values. Yellow samples represents the decrease
in forces during the transition between free and contact motions.
}\label{sampling}
\end{center}
\end{figure}


At any given state fed back by the motion planner (corresponding to the node to be expanded), the reasoning process uses
 geometric reasoning to update the instantiated knowledge with the current information on the manipulation regions and on the control sampling range.
Those manipulation regions  that become useless are deactivated; the others are set active.
A manipulation region becomes useless if it is occupied by an obstacle (i.e. the robot cannot access it), or if the motion of the object is not possible when the robot interacts with it from the manipulation region, e.g. if the front manipulation region of the car-like object is blocked with a \textit{mObject} it is deactivated, as well as the rear \textit{mRegion} because the robot can not exert forces from there until the front \textit{mRegion} becomes free. 
\subsection{Reasoning Process}\label{sec:reasoning}
\begin{figure}[t]
\begin{center}
   \includegraphics[width=0.8 \linewidth]{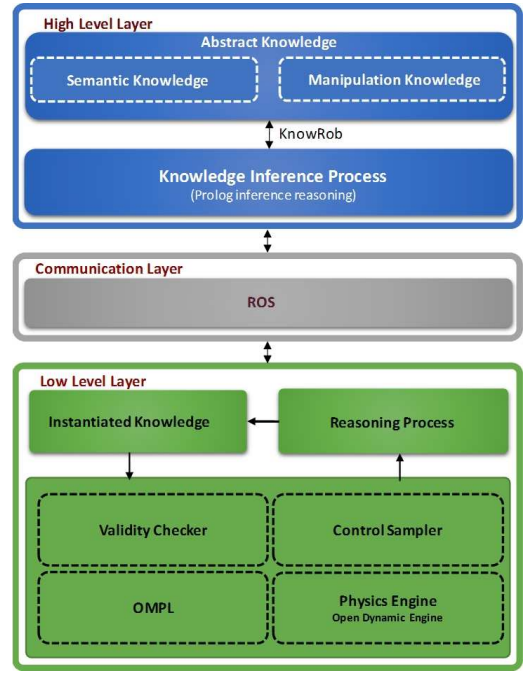}
   \caption{Knowledge-oriented physics-based motion planning (\textit{$\kappa$-PMP}) for power-efficient motion plan.}\label{fig:framwork}
\end{center}
\end{figure}
Kinodynamic planners sample both the direction and module of the control to be applied to extend the tree data structure. The proposed reasoning process computes the module range from where to sample depending on whether the robot  is located in \Cmove\ or \Cint, and in this latter case depending on whether the robot is in collision or not.
The normal control range to move the robot  freely in \Cmove\ will be diminished when being in \Cint, and if contact occurs then it will be increased based on the weight of the object to be pushed.
Let $F$ be the module range from where to sample, and let $F$ take the value $F=[f_\textrm{min},f_\textrm{max}]$ when the robot is in \Cmove. Then, when the robot is in \Cint:
\begin{itemize}
\item $F=\alpha[f_\textrm{min},f_\textrm{max}]$ with $\alpha<1$ if no contact occurs,
\item $F=[f_\textrm{min}+f_\textrm{obj},f_\textrm{max}+f_\textrm{obj}]$ with $f_\textrm{obj}= \mu_\textrm{obj}\; m_\textrm{obj}\; g,$ if contact occurs,
\end{itemize}
where $\mu_\textrm{obj}$ is the friction coefficient between the object and the floor, $m_\textrm{obj}$ is the mass of the object, and $g$ the gravitational force.
In case of a manipulator (kinematic chain), the range of forces is converted to a range of joint torques using the transposed Jacobian.

To illustrate this, Fig.~\ref{sampling} qualitatively depicts the difference between the use of the proposed control sampling range and
 a fixed range, as done in standard physics-based motion planning approaches~\cite{muhayyuddin2015, sucan2012}. In our proposal, the higher value of control forces will only be used when the robot is in contact with the object, and according to its weight. If a fixed lower control range is set then it may not be able to push the objects (to clear the regions) and fails to compute the path, on the other hand, if a higher control range is set then it consumes unnecessary power and may result in a huge displacement of the object. Moreover, prior to contact, the proposed controls slow down to have a smooth transition from no contact to contact.

\section{ The \textit{$\kappa$-PMP} approach}\label{s-imlementation}
\subsection{Proposed Framework}\label{s-framework}

The proposed framework for power-efficient physics-based motion planning is depicted in Fig.~\ref{fig:framwork}. It is a hybrid planning framework that consists of three main layers: the high-level layer devoted to the knowledge management and inference process, the communication layer, and the low-level layer devoted to the physics-based motion planning.

The high-level layer that contains the high-level representation of knowledge is divided into the semantic knowledge coded in the form of ontologies, and the manipulation knowledge inferred from the semantic knowledge using the knowledge inference process module.

The low-level layer consists of the instantiated knowledge~$\kappa$ (that contains the temporary manipulation constraints and the bounds on the control forces), the reasoning process (responsible of updating the instantiated knowledge from the current state and the manipulation knowledge), and the motion planner (whose main modules are the physics-based state validity checker~$\mathcal{F}$, the control sampling module, a standard kinodynamic planner like KPIECE or RRT from the Open Motion Planning Library~\cite{sucanMK2012}, and the ODE-based physics engine). This layer is developed within \textit{The Kautham Project}~\cite{Rosell2014}, an open source framework for motion planning that includes geometric, differential and physics-based motion planners (including those that require ontological knowledge).

The communication layer is based on ROS and communicates the high-level abstract knowledge and the low-level motion planning layer: the abstract knowledge is encapsulated as a ROS service and the motion planning layer accesses it as a ROS client.
\begin{figure*}[h]
\begin{center}
   \includegraphics[width=\textwidth]{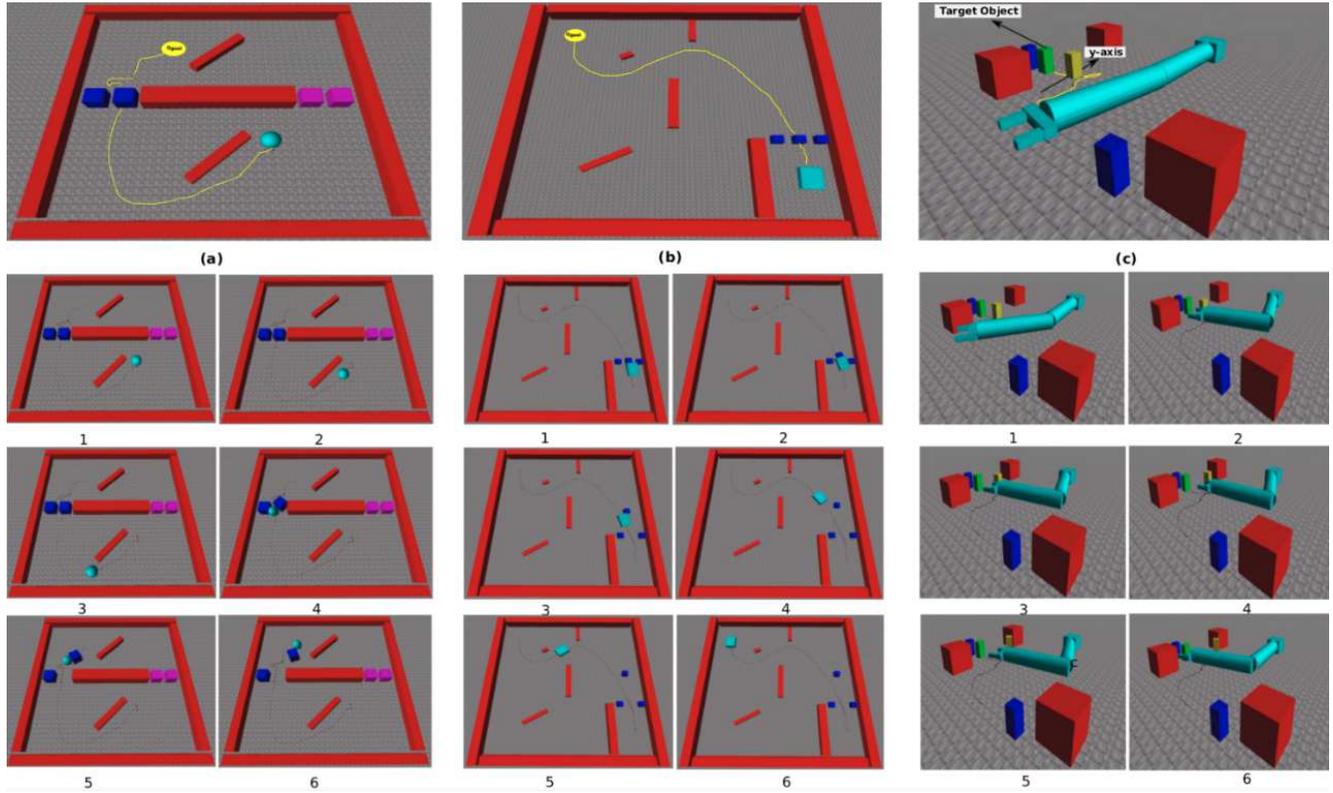}
   \caption{Planning scenes: (a) an holonomic mobile robot; (b) a car-like mobile robot; (c) a planar kinamatic chain. Video: 
https://sir.upc.edu/projects/kautham/videos/k-PMP1.mp4}\label{fig:scenes}
\end{center}
\end{figure*}


\subsection{Algorithm}
\begin{algorithm}[t]
\begin{algorithmic}[1]
{\small
\REQUIRE Initial state $\mibf{q}_{\textrm{init}}$, Goal region $\mibf{Q}_{\textrm{goal}}\in {\cal C}$, Threshold $T_{max}$\\
\ENSURE A path from $\mibf{q}_{\textrm{init}}$ to $ \mibf{q} \in \mibf{Q}_{\textrm{goal}}$
\STATE WorkspaceInit()
\STATE $\mathcal{K}_\textrm{S} \leftarrow$SemanticKnowledgeGenerator()
\STATE $\mathcal{K}_\textrm{M} \leftarrow$ManipulationKnowledgeInference($\mathcal{K}_\textrm{S}$)
\WHILE {$t<T_{max}$}
\STATE  $\kappa \leftarrow$ ReasoningProcess($\mathcal{K}_\textrm{M}, \mathbf{q}$)
\STATE SelectNodeToExpand()
\STATE $\{u,n\}$ $\leftarrow$ SampleControlsAndSteps($\kappa$)
\FOR{$i=0$ \TO $i<n$ }
\STATE $\mibf{q}_{\textrm{new}}\leftarrow$ Propagate($\mibf{q},u$)
\IF{! StateValidityChecker($\mibf{q}_{\textrm{new}}$,$\kappa$)}
\STATE Break
\ELSE
\STATE UpdateConnections()
\ENDIF
\ENDFOR
\IF {$\mibf{q}_{\textrm{new}} \in \mibf{Q}_{\textrm{goal}}$}
           \RETURN Path($\mibf{q}_{\textrm{new}}$)
\ENDIF
\ENDWHILE
\RETURN {\small \sf NULL}
}
\end{algorithmic}
\caption{$\kappa\textrm{-}PMP$}
\label{alg:Algorithm1}
\end{algorithm}
\begin{algorithm}[t]
\begin{algorithmic}[1]
{\small
 \STATE $\Gamma\leftarrow$ UpdateManipulationConstraints($\mathcal{K}_M,\mathbf{q}$)
 \STATE $\mathcal{L}\leftarrow$ ComputeRobotLocation($\mathbf{q}$)
 \STATE $\{f_\textrm{min},f_\textrm{max}\}\leftarrow$ComputeControlRange($\mathcal{L}$)
 \STATE $\kappa_\textrm{new}\leftarrow$UpdateInstantiatedKnowledge($\Gamma,\{f_\textrm{min},f_\textrm{max}\}$)
 \RETURN {$\kappa_\textrm{new}$}
}
\end{algorithmic}
\caption{ReasoningProcess($\mathcal{K}_\textrm{M}, \mathbf{q}$)}
\label{alg:Algorithm2}
\end{algorithm}

The planning process is sketched in algorithm~\ref{alg:Algorithm1}. It takes as input the initial state $\mathbf{q}_{init}$, the goal region $\mathbf{Q}_{goal}$, and the maximum allowed planning time $T_{max}$. If a solution is found, it returns the path from $\mathbf{q}_{init}$ to $\mathbf{q}_{goal}\in\mathbf{Q}_{goal}$, or NULL otherwise. To sample the states and construct the planner  data structure, the approach provides the flexibility to use any sampling-based kinodynamic motion planner (such as RRT, KPIECE, SyCLoP) offered by OMPL, together with the Open Dynamic Engine as state propagator. 

Lines~[1-3] of the algorithm are the preprocessing steps for the motion planner; lines~[4-15] contain the planning process. The main functions used are the following:
\begin{itemize}
 \item \textit{WorkspaceInit:} Iinitializes the state of the robot and of the objects in the environment.
 \item \textit{SemanticKnowledgeGenerator:} Creates the semantic knowledge $\mathcal{K}_S$ from the ontologies. 
 \item \textit{ManipulationKnowledgeInference:} Infers $\mathcal{K}_M$ from $\mathcal{K}_S$. 
 \item \textit{ReasoningProcess:} Updates the instantiated knowledge by updating the manipulation constraints and the range of the control, as detailed in algorithm~\ref{alg:Algorithm2}.
\item \textit{SelectNodeToExpand:} Selects the node to expand following the node selection process of the kinodynamic planner used.
\item \textit{SampleControlsAndSteps:} Samples the controls and the number of steps describing the number of times the selected control will be applied repeatedly. To sample the controls any control 
sampling strategy, such as steering control sampling, can be used.
\item \textit{Propagate:} Applies the sampled control $\mathbf{u}$ on the state $\mathbf{q}$ for $\Delta t$ time using the Open Dynamic Engine as state propagator.
\item \textit{StateValidityChecker}: Is the physics-based state validity checker ($\mathcal{F}$) that validates the newly generated state by taking into account the instantiated knowledge.
\item \textit{UpdateConnections:} Adds the accepted state to the planner data structure and updates the connections accordingly.
\item \textit{Path:} Returns a path from $\mathbf{q}_{init}$ to $\mathbf{q}_{goal}$ if the last generated state lies in the goal region, and NULL otherwise.
\end{itemize}
 
Algorithm~\ref{alg:Algorithm2} implements the reasoning process, it contains the following steps:
\begin{itemize}
 \item \textit{UpdateManipulationRegions:} Activates and deactivates the manipulation regions 
 using the geometric reasoning based on $\mathcal{K}_M$ and the current state. 
 \item \textit{ComputeRobotLocation:} Determines the region $\mathcal{L}$ \\where the robot lies.
 \item \textit{ComputeControlRange:} Determines the control range as a function of~$\mathcal{L}$.
 \item \textit{UpdateInstantiatedKnowledge:} Updates the data \\structures of the instantiated knowledge with the updated manipulation regions and the control sampling range.
\end{itemize}
The key point of the algorithm is that, during planning,  the instantiated knowledge is updated at each instant of time by the reasoning process (Line~5), conditioning the sampling of the 
controls (Line~7) and the state validity checking procedure (Line~9).
\begin{figure}[h]
\begin{center}
   \includegraphics[width=1.0\linewidth]{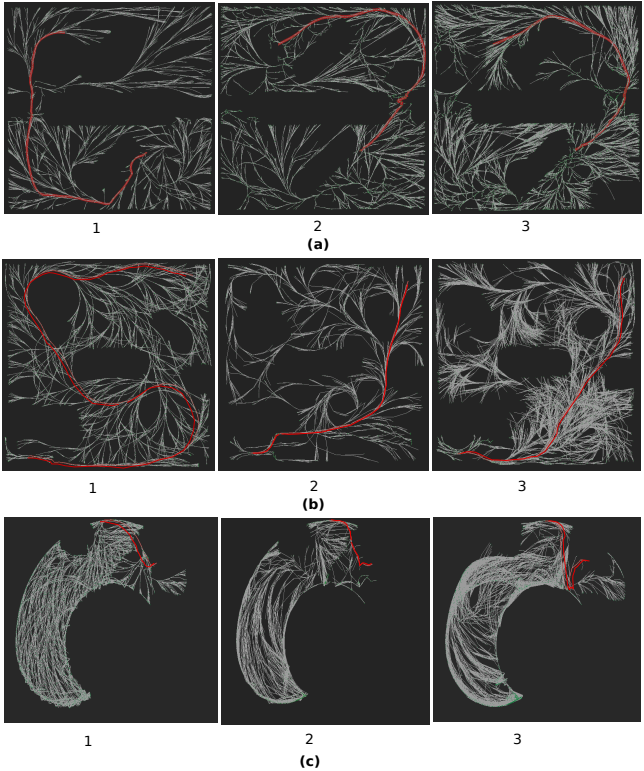}
   \caption{State space projection onto the workspace for the scenes with the holonomic mobile robot (top), the car-like robot (middle) and the planar manipulator (bottom) using: (1) \textit{$\kappa$-KPIECE}, (2) 
\textit{$\kappa$-RRT} and (3) \textit{$\kappa$-SyCLoP} planners.}\label{fig:cspace}
\end{center}
\end{figure}
\subsection{Simulation Setups}
The proposed approach is validated using three different robot models, depicted in Fig.~\ref{fig:scenes}. The scenario presented in Fig.~\ref{fig:scenes}-a is for an holonomic mobile robot. It consists of a robot (sphere), \textit{free-mObjects} (blue and purple cubes, being the purple ones heavier than blue cubes) and \textit{fixed-objects} (red walls). The problem is to go from $\mathbf{q}_{init}$ to $\mathbf{q}_{goal}$, being the possible solutions blocked by the blue or the purple cubes. Since the motion of this robot is controlled by applying the control force, the range of the control force will be selected by the reasoning process according to the object to be pushed (blue or purple), as explained in Sec.~\ref{sec:reasoning}. The higher values of controls will only be applied when the robot is in interaction with the object, unlike the conventional planners that fix the control range at the start, our approach dynamically varies the control range. The power $\mathcal{P}$ consumed by the robot while moving along the path is computed as

 \begin{equation}
\label{eq:power}
  \mathcal{P}  =\sum\limits_{i}^n \frac{ \mathbf{f}_i\cdot\mathbf{d}_i}{\Delta t_i},
\end{equation}
with $\mathbf{f}$ and $\mathbf{d}$ being, respectively, the applied force and displacement vectors, and $\Delta t$ the time duration.

\begin{figure*}[t]
\begin{center}
   \includegraphics[width=0.85\linewidth]{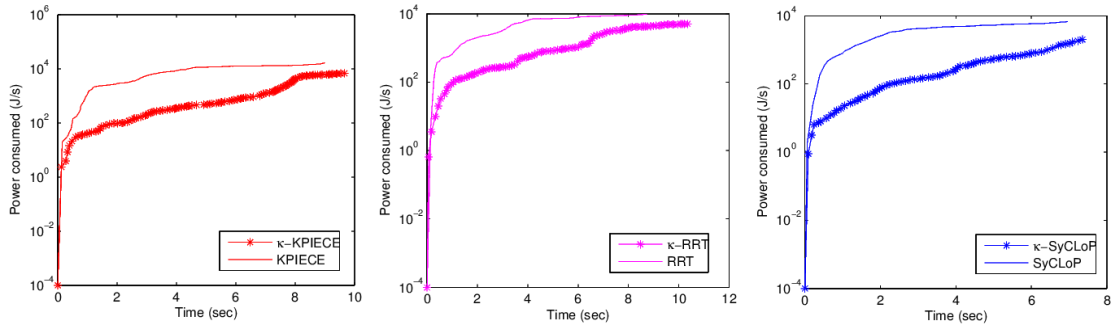}
   \caption{Logarithmic plot of the power consumed while moving along the path for the holonomic mobile robot.}\label{fig:power1}
\end{center}
\end{figure*}
\begin{figure*}[t]
\begin{center}
   \includegraphics[width=0.85\linewidth]{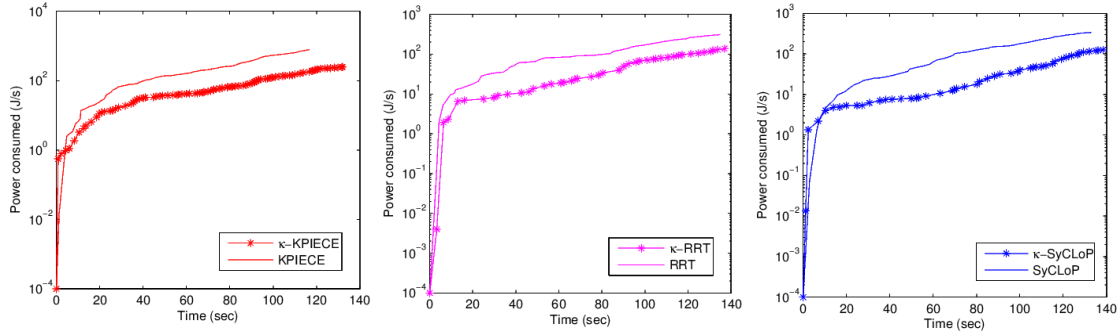}
   \caption{Logarithmic plot of the power consumed while moving along the path for the car-like robot.}\label{fig:power2}
\end{center}
\end{figure*}
\begin{figure*}[t]
\begin{center}
   \includegraphics[width=0.85\linewidth]{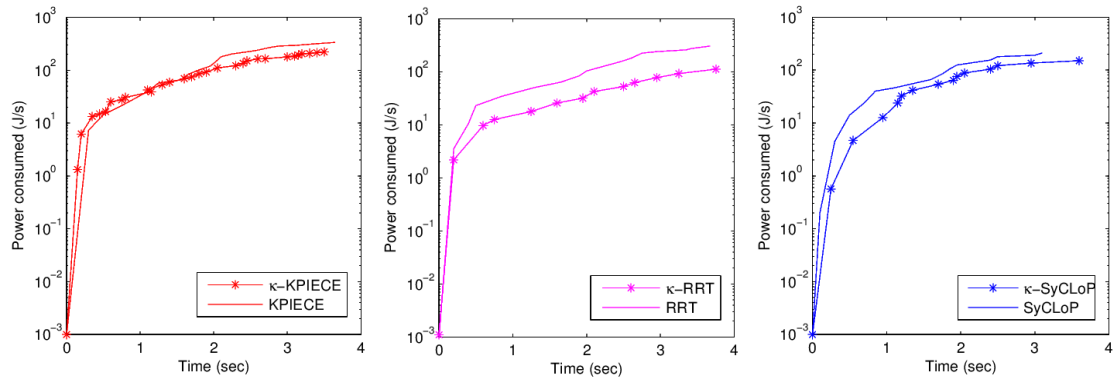}
   \caption{Logarithmic plot of the power consumed while moving along the path for the planar manipulator.}\label{fig:power3}
\end{center}
\end{figure*}
Fig.\ref{fig:scenes}-b describes the scene for the car-like robot. The position and orientation of the car is controlled by adding the torque to the wheels and to the steering. The path to the goal is blocked with the \textit{free-mObjects} (blue cubes) and in order to reach the goal the car has to clear the way by pushing the objects away. The high amount of torque is only required while pushing the objects. The reasoning process updates the torque bounds by determining the forces that are required to push the object, and transforms it into the torque exerted by each wheel. The power consumed while moving along the path is computed as
\begin{equation}
\label{eq:rotpower}
  \mathcal{P}  =\sum\limits_{i}^n \mathbf{\tau}_i\cdot\mathbf{\omega}_i,
\end{equation}
where $\tau$ is the torque exerted by the wheels and $\omega$ is the corresponding angular velocity. 

The scenario for the planar manipulator is depicted in Fig.~\ref{fig:scenes}-c, it consists of \textit{free-mObject} 
(yellow and blue boxes), the target object (green box) and the \textit{fixedObjects} (red cubes). The manipulator is shown in its initial state and the goal is to grasp the target object, being the way to that object  blocked by the yellow box. Since one region of the yellow box is occupied with the target object, the reasoning process will change its type to \textit{co-mObject} and it can only be manipulate along y-axis (Once all the manipulation regions of the yellow box will be free, its type will again update to \textit{free-mObject}). In-order to reach the goal the manipulator has to push it away.  The control forces are transformed into joint torques using the 
transposed Jacobian and the power is computed as
\begin{figure}[h]
\begin{center}
   \includegraphics[width=.85\linewidth]{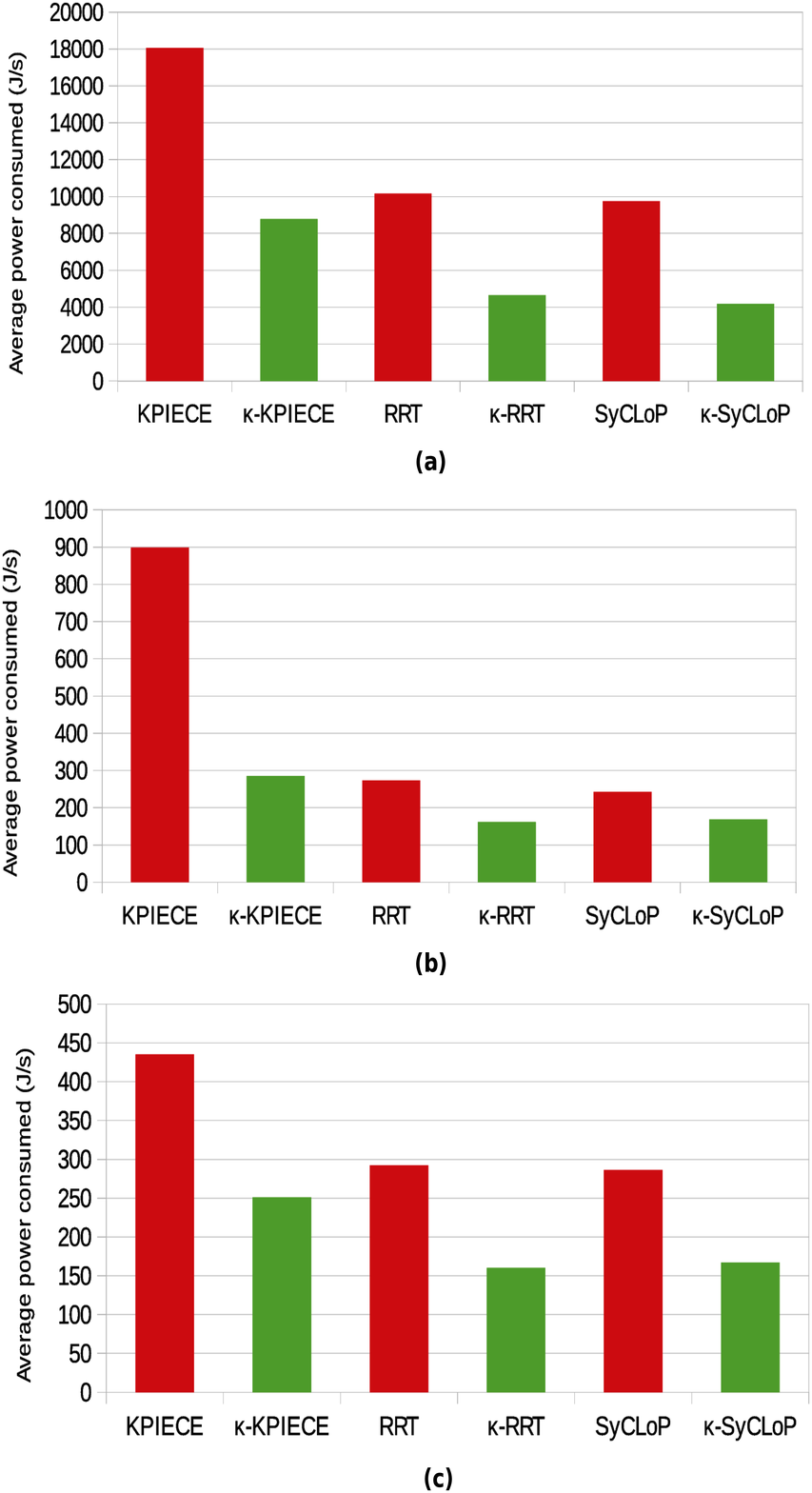}
   \caption{Histogram of the average power consumed by the robots: (a) holonomic mobile robot, (b) car-like robot, (c) planar manipulator.}\label{fig:phistogram}
\end{center}
\end{figure}
\begin{figure}[h]
\begin{center}
   \includegraphics[width=0.85\linewidth]{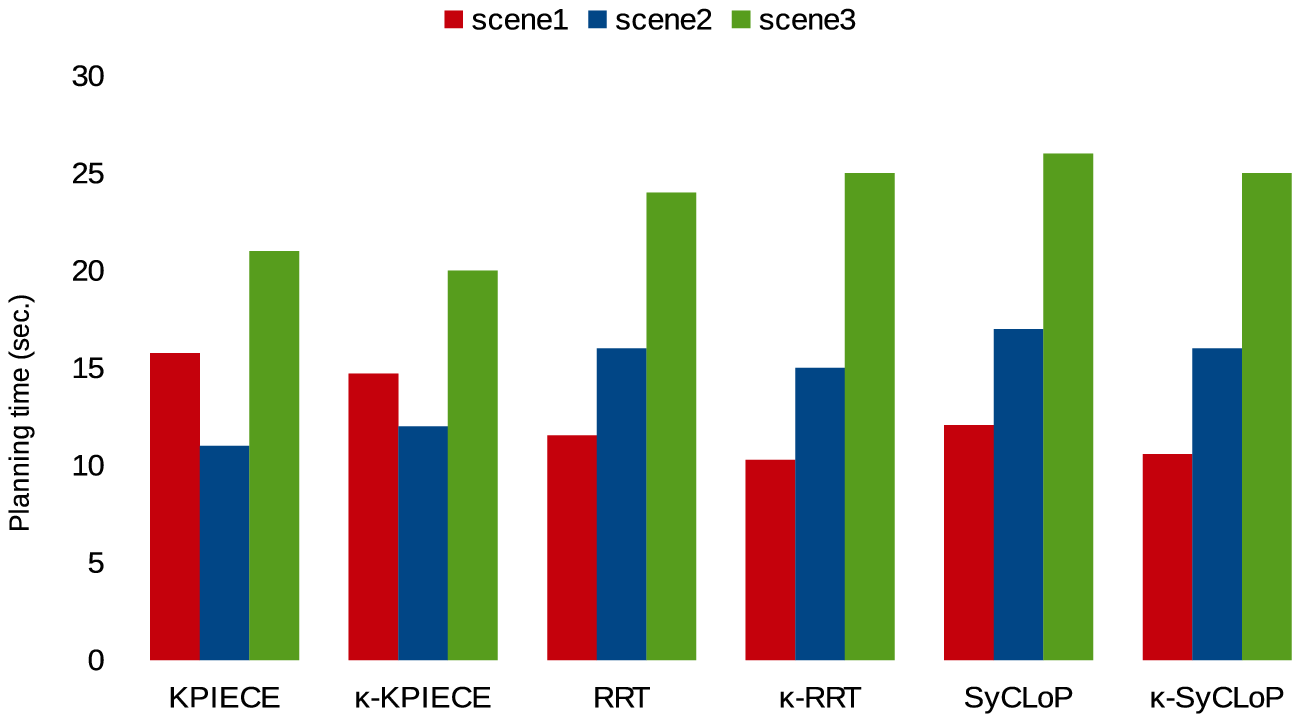}
   \caption{Histogram of the average planning time. Scene1, scene2 and scene3 correspond to the scenes illustrated in Fig. \ref{fig:scenes}.}\label{fig:planningtime}
   \end{center}
\end{figure}
\begin{equation}
\label{eq:mpower}
  \mathcal{P}  =\sum\limits_{i}^n \mathbf{\tau}_i\cdot\mathbf{\omega}_i,
\end{equation}
where now $\tau$ is the torque exerted by the joints and $\omega$ is the corresponding angular joint velocity. Fig.~\ref{fig:cspace} depicts the configuration space for the \textit{$\kappa$-KPIECE}, 
\textit{$\kappa$-RRT} and \textit{$\kappa$-SyCLoP} planners for the above stated scenarios.

\begin{figure}[h]
\begin{center}
   \includegraphics[width=0.85\linewidth]{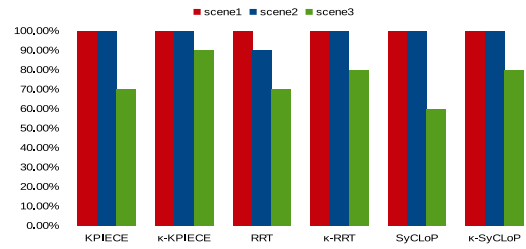}
   \caption{Histogram of the success rate. Scene1, scene2 and scene3 correspond to the scenes illustrated in Fig. \ref{fig:scenes}}\label{fig:successrate}
\end{center}
\end{figure}

\section{Results and Discussion}\label{s-resultsdiscussion}

 We compared \textit{$\kappa$-PMP} with simple physics-based planning, using RRT, KPIECE and SyCLoP as kinodynamic motion planners, because a recent benchmarking study~\cite{gillani2016} of the physics-based motion planning showed these planners as being the most suitable for physics-based motion planning: KPIECE computed the efficient solutions in terms of time whereas SyCLoP and RRT computed the efficient solution in terms of power.
No comparison has been done with other planners that seek different goals, such as ~\cite{Haustein2015,stilman2005} that cope with the rearrangement of the objects in the workspace or such as \cite{li2016} that are focus on optimization issues. 

The parameter used in the comparison  were: 

\begin{itemize}
\item \textit{Power consumption:} The total power consumed by the robot while moving along the solution path.
\item \textit{Planning time:} The total time consumed by the planner to compute the solution path.
\item \textit{Success rate:} The number of successful runs in the maximum limit of planning time.
\end{itemize}
\begin{figure}[h]
\begin{center}
   \includegraphics[width=1\linewidth]{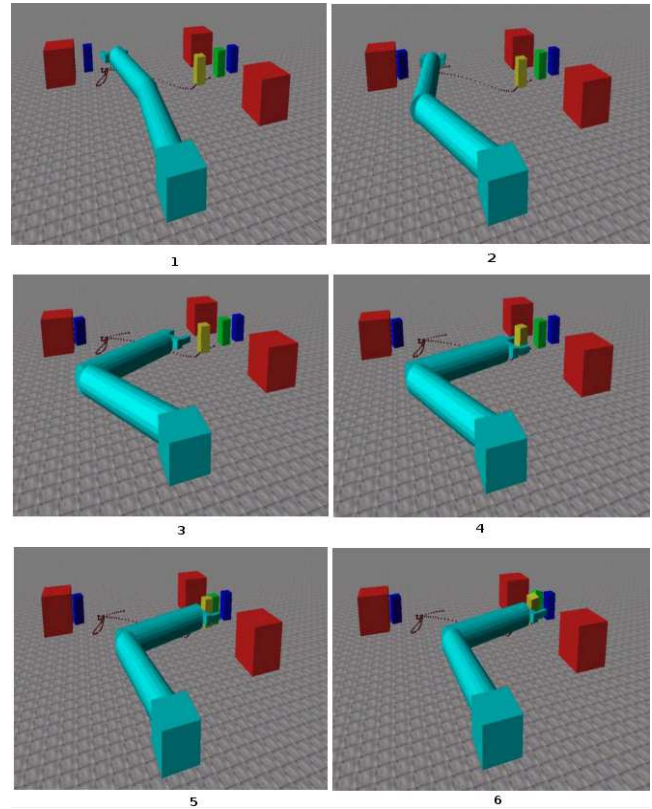}
   \caption{Execution of the computed motion plan by the simple physics-based motion planner. It can be observed that the task fails because the manipulatable yellow object ends at the gripper, thus 
preventing the manipulator to grasp the target green object.}\label{fig:wrong}
\end{center}
\end{figure}
\begin{figure}[h]
\begin{center}
   \includegraphics[width=1\linewidth]{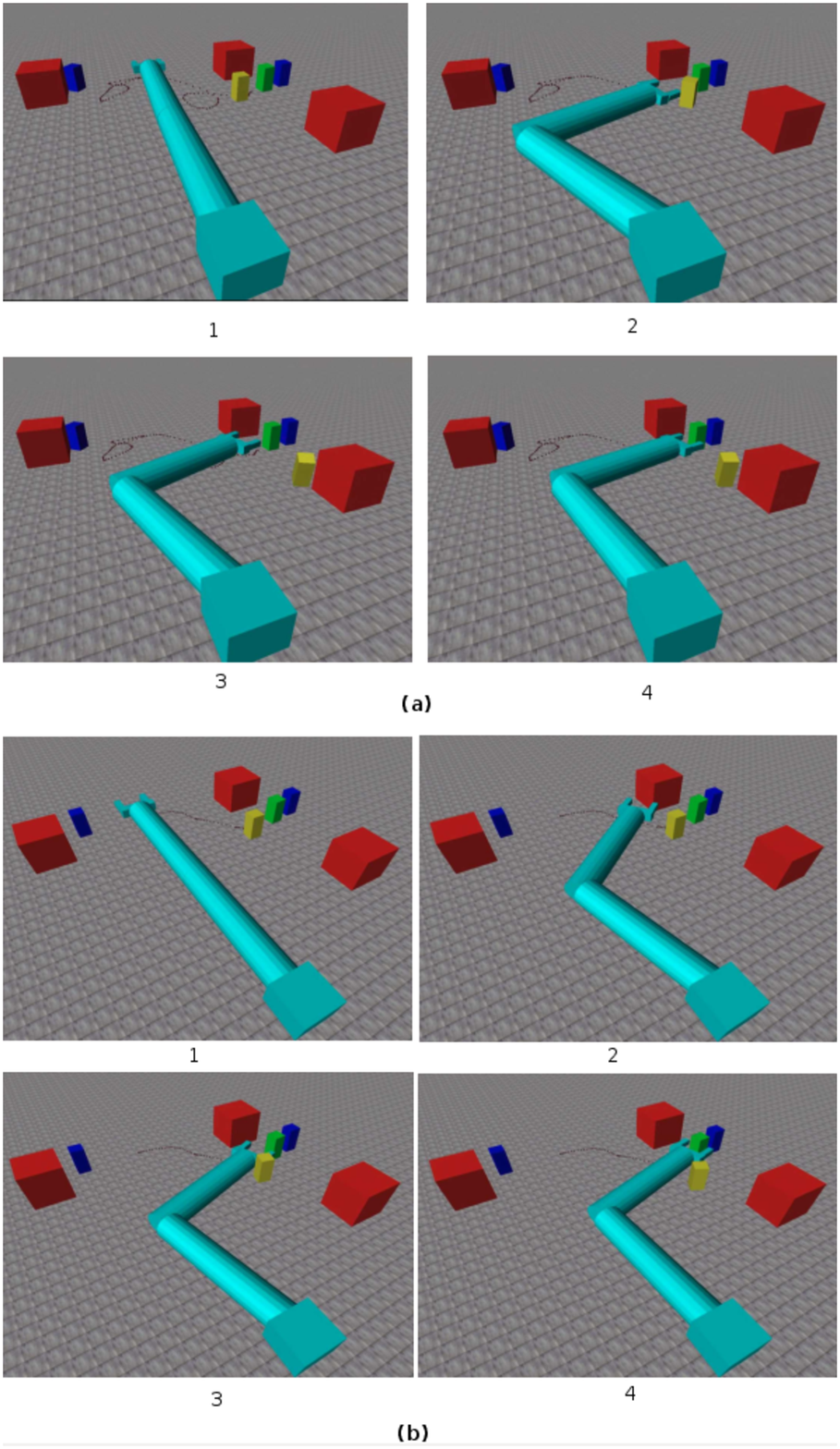}
   \caption{ Snapshots of the dynamic interactions between the robot and the yellow object corresponding to the  simple physics-based motion planning (a) and to 
the \textit{$\kappa$-PMP} planner (b), respectively. Videos: https://sir.upc.edu/projects/kautham/videos/k-PMP2.mp4}\label{fig:push}
\end{center}
\end{figure}
\subsection{Quantitative Analysis}
\textit{$\kappa$-PMP} computes the power efficient solution as compared to the simple physics-based motion planning approach. We compared the best results that we obtained using both approaches with 
different kinodynamic motion planners. Fig.~\ref{fig:power1}, Fig.~\ref{fig:power2}, and Fig.~\ref{fig:power3} show the logarithmic plot of power consumption vs time corresponding to the holonomic 
mobile robot, the car-like robot and the planar manipulator, respectively. In all cases \textit{$\kappa$-PMP} is the most efficient plan in terms of power.
The average power consumed by each planner in several runs is shown in the form of histogram (Fig.~\ref{fig:phistogram}) corresponding to the three different robot models. There is a significant 
difference in terms of power consumption when using \textit{$\kappa$-PMP} and simple physics-based planning approaches.

The average planing time of several runs for three scenes is presented in Fig.~\ref{fig:planningtime}. \textit{$\kappa$-PMP} computes the solution almost in the same time as the traditional approach 
does. To compute the success rate, we set the maximum planning time to 150 s and each query is executed 10 times. Fig.~\ref{fig:successrate} shows the histogram of the success rate where it can be 
appreciated that  \textit{$\kappa$-PMP} has a similar success rate as the traditional one. In some cases (particularly for the manipulator) our approach has the higher success rate than the 
traditional one.

\subsection{Qualitative Analysis}
The analysis of the results shown in the previous section illustrates that \textit{$\kappa$-PMP} preserves the behavior of the kinodynamic planner used (such as RRT or KPIECE) and significantly reduce 
the power consumed. This is due to the fact that \textit{$\kappa$-PMP} dynamically varies the control sampling range according to the physical properties of the target object. Moreover, 
\textit{$\kappa$-PMP} uses knowledge to determine the way to manipulate objects in order to reach the goal in a more realistic way. For instance, Fig.~\ref{fig:wrong} shows a sequence of snapshots of 
an execution using a traditional physics-based planning approach. Since it lacks the knowledge about the way of manipulating the objects, the yellow box ends  stuck in the gripper, thus preventing the 
manipulator to grasp the green box, reducing the success rate of the planner. 

\textit{$\kappa$-PMP} manipulates the objects in a natural way without putting any extra constraints (such as quasi static push). Fig.~\ref{fig:push} shows the results of the dynamic interaction 
between the robot and the object using both approaches. In most of the cases simple physics-based planning approach threw the object away (Fig.~\ref{fig:push}-a) because it is unaware of how much 
force is required to push. In contrast, Fig.~\ref{fig:push}-b shows the smoother interaction resulting from the use of \textit{$\kappa$-PMP}. It applies the forces based on the physical 
properties (such as mass and friction) and pushes the object in a smooth way. Furthermore the solution path computed by the 
$\kappa$-PMP 
is naturally biased towards the manipulation regions, which helps to effectively manipulate the objects.
\subsection{Practical Application of the Proposed Approach}
From the analysis done, it has been shown that the proposed approach computes more robust and power efficient motion trajectories, as compared to the kinodynamic or simple physics-based planning 
approaches (mentioned 
in the related work). Moreover, the current proposal handles dynamic interaction in a more natural way.  
With a practical perspective, therefore, this approach may be significantly important in two directions. On the one hand, it can be a part of an integrated task and motion planner. The availability of power-efficient motion trajectories may have a relevant contribution in the final performance of the integrated task and motion planner, since  the costs of the actions in a plan are critical and greatly influence  the decisions of the task planner. 
On the other hand, as a stand alone planner, it results in a smart and powerful tool to manipulate objects in the clutter, without the need of reasoning at task level, giving practical solutions to 
quite difficult problems.

\section{Conclusion}\label{s-conclusion}
This paper has proposed a framework, called \textit{$\kappa$-PMP}, to use knowledge to enhance physics-based motion planners based on any kinodynamic algorithm, like RRT or KPIECE, and on any dynamic 
engine, like ODE or Bullet.  Manipulation knowledge, inferred from an abstract knowledge ontology coded using the Web Ontology Language (OWL), is used to incorporate a reasoning process within the 
state transition model. This  allows to dynamically update: a) the control sampling range based on the region of the configuration space where the robot is located and on the physical properties of 
the objects to be interacted; b) the regions from where an object can be manipulated. A comparison of physics-based motion planners based on RRT, KPIECE and SyCLoP with and without using the 
\textit{$\kappa$-PMP} framework has been carried out in three different scenarios involving a holonomic mobile robot, a car-like robot and a planar manipulator. The proposal resulted in an improvement 
of the success rate and of the performance in terms of power consumed and quality of the solutions.

\balance
\bibliographystyle{spmpsci}      
\bibliography{References}
 \vspace{5mm}

 \begin{wrapfigure}{l}{22mm} 
 \vspace{-7mm}
    \includegraphics[width=1in,height=1.25in]{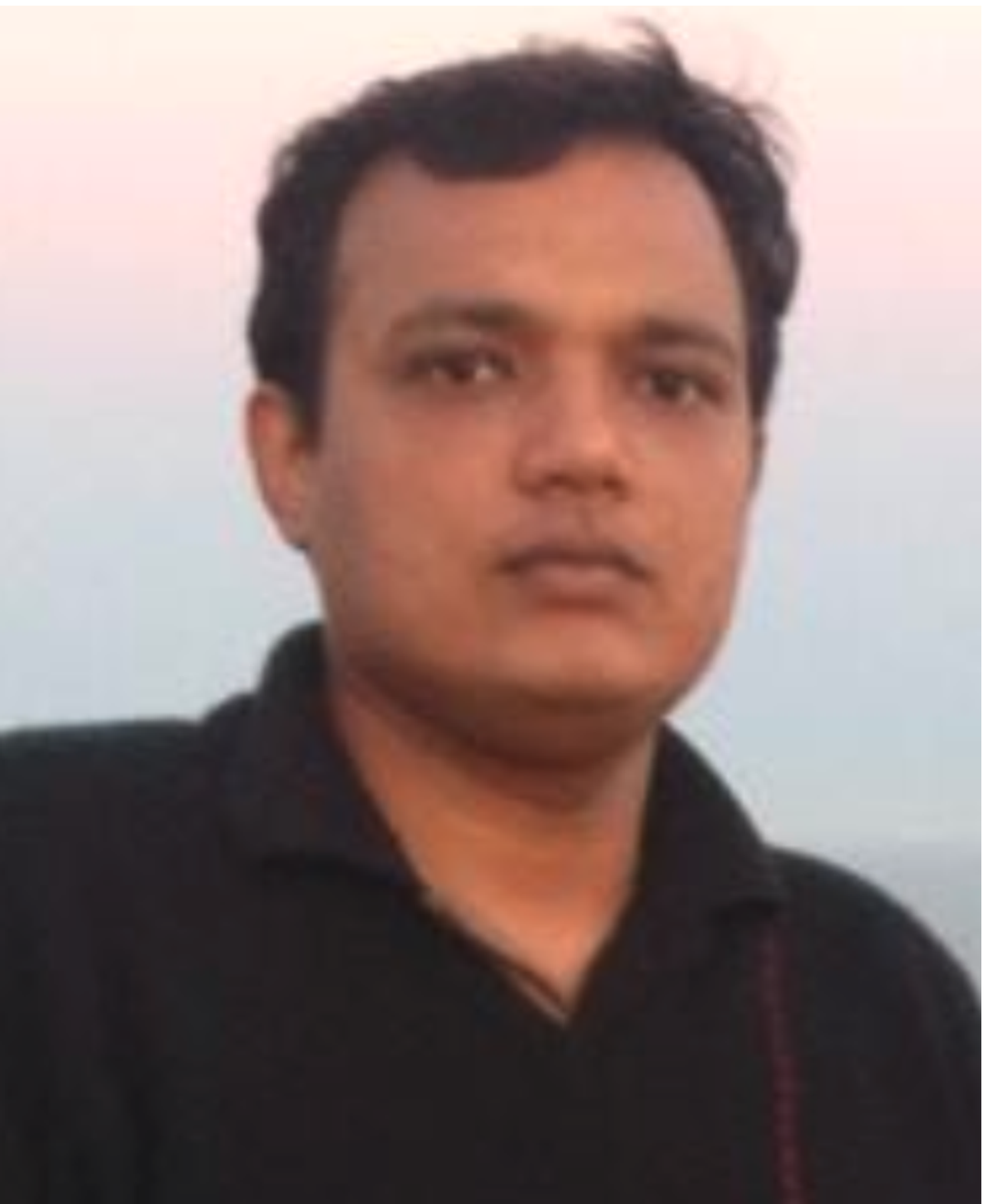}
  \end{wrapfigure}\par
  \textbf{Muhayyuddin}  is a Ph.D. student in Automatic Control, Robotics and Computer Vision, at the Institute of Industrial and Control Engineering, Universitat Polit\`ecnica de Catalunya
(UPC), Barcelona Spain. His current research area includes physics-based motion planning for grasping and manipulation, dynamic simulations and mobile manipulation. He Received the BS degree in 
Computational Physics from University of the Punjab, Lahore, Pakistan in 2008 and the MS degree in Computer Science from GC University Lahore, Pakistan in 2011.
 \vspace{5mm}

 \begin{wrapfigure}{l}{22mm} 
 \vspace{-7mm}
    \includegraphics[width=1in,height=1.25in]{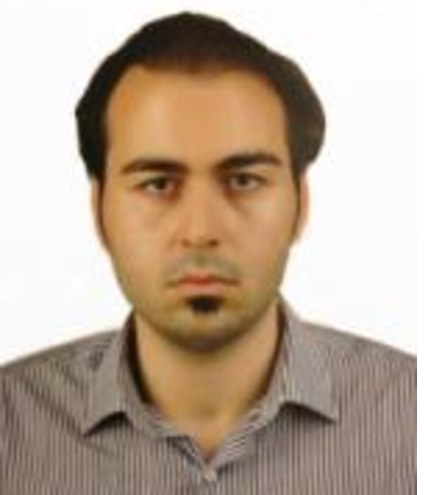}
  \end{wrapfigure}\par
  \textbf{ Aliakbar Akbari} received the B.Sc. in Mechanical Engineering at the Islamic Azad University Parand Branch and the M.Sc. degree in Mechanical Engineering at the Eastern 
Mediterranean University (EMU). He is a PhD.  student in Automatic Control, Robotics and Computer Vision at the Institute of Industrial and Control Engineering, Universitat Polit\`ecnica de Catalunya 
(UPC). His current research areas include combination of knowledge-based task and physics-based motion planning and mobile manipulation.
 \vspace{5mm}

\begin{wrapfigure}{l}{22mm} 
 \vspace{-7mm}
    \includegraphics[width=1in,height=1.25in]{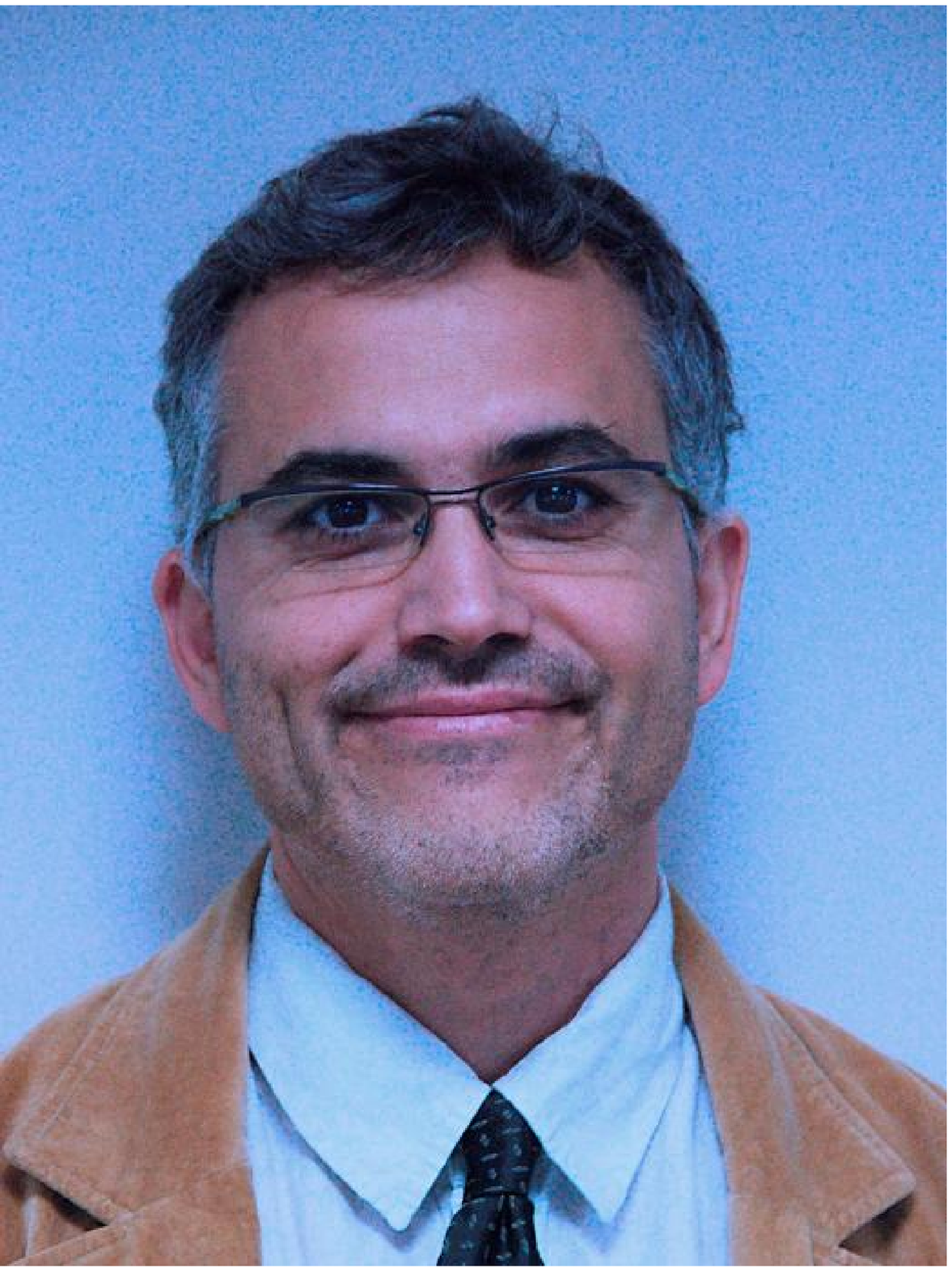}
  \end{wrapfigure}\par
  \textbf{ Jan Rosell}  received the BS degree in Telecommunication Engineering and the Ph.D. degree in Advanced Automation and Robotics from the Universitat Polit\`ecnica de Catalunya (UPC), 
Barcelona, Spain, in 1989 and 1998, respectively. He joined the Institute of Industrial and Control Engineering (IOC) in 1992 where he has developed research activities in robotics. He has been 
involved in teaching activities in Automatic Control and Robotics as Assistant Professor since 1996 and as Associate Professor since 2001. His current technical areas include task and motion planning, 
mobile manipulation, and robot co-workers.

\end{document}